\documentclass[sigconf]{acmart}
\usepackage{subcaption}
\AtBeginDocument{%
  \providecommand\BibTeX{{%
    \normalfont B\kern-0.5em{\scshape i\kern-0.25em b}\kern-0.8em\TeX}}}

\copyrightyear{2020}
\acmYear{2020}
\setcopyright{acmcopyright}
\acmConference[FAT* '20]{Conference on Fairness, Accountability, and Transparency}{January 27--30, 2020}{Barcelona, Spain}
\acmBooktitle{Conference on Fairness, Accountability, and Transparency (FAT* '20), January 27--30, 2020, Barcelona, Spain}
\acmPrice{15.00}
\acmDOI{10.1145/3351095.3372852}
\acmISBN{978-1-4503-6936-7/20/02}

\begin{document}

\title{Effect of Confidence and Explanation on Accuracy and Trust Calibration in AI-Assisted Decision Making}


\author{Yunfeng Zhang}
\email{zhangyun@us.ibm.com}
\affiliation{%
  \institution{IBM Research AI}
  \streetaddress{1101 Kitchawan Road}
  \city{Yorktown Heights}
  \country{USA}}
\author{Q. Vera Liao}
\email{vera.liao@ibm.com}
\affiliation{%
  \institution{IBM Research AI}
  \streetaddress{1101 Kitchawan Road}
  \city{Yorktown Heights}
  \country{USA}}
\author{Rachel K. E. Bellamy}
\email{rachel@us.ibm.com}
\affiliation{%
  \institution{IBM Research AI}
  \streetaddress{1101 Kitchawan Road}
  \city{Yorktown Heights}
  \country{USA}}


\begin{abstract}

Today, AI is being increasingly used to help human experts make decisions in high-stakes scenarios. In these scenarios, full automation is often undesirable, not only due to the significance of the outcome, but also because human experts can draw on their domain knowledge complementary to the model's to ensure task success. We refer to these scenarios as AI-assisted decision making, where the individual strengths of the human and the AI come together to optimize the joint decision outcome. A key to their success is to appropriately \textit{calibrate} human trust in the AI on a case-by-case basis; knowing when to trust or distrust the AI allows the human expert to appropriately apply their knowledge, improving decision outcomes in cases where the model is likely to perform poorly. This research conducts a case study of AI-assisted decision making in which humans and AI have comparable performance alone, and explores whether features that reveal case-specific model information can calibrate trust and improve the joint performance of the human and AI. Specifically, we study the effect of showing confidence score and local explanation for a particular prediction. Through two human experiments, we show that confidence score can help calibrate people's trust in an AI model, but trust calibration alone is not sufficient to improve AI-assisted decision making, which may also depend on whether the human can bring in enough unique knowledge to complement the AI's errors. We also highlight the problems in using local explanation for AI-assisted decision making scenarios and invite the research community to explore new approaches to explainability for calibrating human trust in AI.

\end{abstract}




\maketitle

\section{Introduction}
 Artificial Intelligence (AI) technologies, especially Machine Learning (ML), have become ubiquitous and are increasingly used in a wide range of tasks. While algorithms can perform impressively, in many situations full delegation to ML models is not desired because their probabilistic nature means that there is never a guarantee of correctness for a particular decision. Furthermore, ML models are only as accurate as the historical data used to train them, and this data could suffer from input error, unknown flaws, and biases. ML models can assist human decision-makers to produce a \textit{joint decision outcome} that is hopefully better than what could be produced by either the model or human alone. Ultimately however, humans would be responsible for the decisions made. Therefore ML decision-support applications should be developed not only with the goal of high performance, safety and fairness, but also allowing the decision-maker to understand the predictions made by the model. This is especially important for decision-making in high-stakes situations affecting human lives such as medical diagnosis, law enforcement, and financial investment. 

A key to success in AI-assisted decision making is to form a correct mental model of the model's error boundaries~\cite{bansal2019updates}. That is, the decision-makers need to know when to trust or distrust the model's recommendations. If they mistakenly follow the model's recommendations at times when it is likely to err, the decision outcome would suffer, and catastrophic failures could happen in high-stakes decisions. Many have called out the challenges for humans to form a clear mental model of an AI, since opaque, "black-box" models are increasingly used. Furthermore, by exclusively focusing on optimizing model performance, developers of AI systems often neglect the system users' needs for developing a good mental model of the AI's error boundaries. For example, frequently updating the AI algorithm may cause confusion to the human decision-maker, who may accept or reject the AI's recommendations at a wrong time, even if the algorithm's overall performance improved \cite{bansal2019updates}. 

To help people develop a mental model of an ML model's error boundaries means to correctly \textit{calibrate} trust on a case-by-case basis. We emphasize that this goal is distinct from \textit{enhancing} trust in AI. For example, while research repeatedly demonstrates that providing high-performance indicators of an AI system, such as showing high accuracy scores, could enhance people's trust and acceptance of the system \cite{Lai2019a,Yu2019,Yin2019}, they may not help people distinguish cases they can trust from those they should not. Meanwhile, ML is probabilistic and the probability of each single prediction can be indicated by a \textit{confidence score}. In other words, the confidence scores reflect the chances that the AI is correct. Therefore, to optimize for the joint decisions, in theory people should rely on the AI in cases where it has high confidence, and use their own judgment in cases where it has low confidence. However, in practice, we know little about how confidence scores are perceived by people, or how they impact human trust and actions in AI-assisted decisions.

To improve people's distrust in ML models, many considered the importance of transparency by providing explanations for the ML model~\cite{Ribeiro2016, Doshi-Velez2017,Carvalho2019}. In particular, local explanations that explain the rationale for a single prediction (in contrast to global explanations describing the overall logic of the model) are recommended to help people judge whether to trust a model on a case-by-case basis~\cite{Ribeiro2016}. For example, many local explanation techniques explain a prediction by how each attribute of the case contributes to the model's prediction~\cite{Ribeiro2016,NIPS2017_7062}. It is possible that in low-certainty cases none of the features stands out to make strong contributions. So the explanation may appear ambivalent, thus alarming people to distrust the prediction. While such a motivation to help people calibrate trust underlies the development of local explanation techniques, to the best of our knowledge, this assumption has not been empirically tested in the context of AI-assisted decision making.

In this paper, we conduct a case study of AI-assisted decision-making and examine the impact of information designs that reveal case-specific model information, including confidence score and local explanation, on people's trust in the AI and the decision outcome. We explored two types of AI-assisted decision-making scenarios. One where the AI gave direct recommendation, and the other where the decision-maker had to choose whether to delegate the decision without seeing the AI's prediction, the latter of which represents a stricter test of trust. We designed the study in a way to have the human decision-makers performing comparably to the AI, and also explored a situation where the humans know they had more domain knowledge than the AI. In contrast, prior works studying AI-assisted decision-making often used setups where humans' decision performance was significantly inferior than the model's~\cite{Lai2019a,Yin2019}, which would by default reward people to rely on the AI. While such a setup is appropriate for studying how to enhance trust in AI, our focus is to study the calibration of trust for cases where the AI has high or low certainty. This paper makes the following contribution:
\begin{itemize}
\item We highlight the problem of \textbf{trust calibration} in AI at a prediction specific level, which is especially important to the success of AI-assisted decision-making.
\item We demonstrate that showing prediction specific \textbf{confidence information} could support trust calibration, even in situations where the human has to blindly delegate the decision to the AI. However, whether trust calibration could translate into improved joint decision outcome may depend on other factors, such as whether the human can bring in a unique set of knowledge that complements the AI's errors. We consider the concept of \textit{error boundary alignment} between the human and the AI, and its implication for studying different AI-assisted decision making scenarios.
\item We show that local, prediction specific \textbf{explanations} may not be able to create a perceivable effect for trust calibration, even though they were theoretically motivated for such tasks. We discuss the limitations of the explanation design we used, and future directions for developing explanations that can better support trust calibration. 

\end{itemize}

\section{Related Work}

The concept of trust has its roots in relationships between humans, reflected in many aspects of collaborative behaviors with others such as willingness to depend, give information and make purchase~\cite{HarrisonMcKnight1998,McKnight2002}. Trust has been widely studied in human-computer and human-machine interaction since users' decisions to continue using a system or accept output from a machine are highly trust-dependant behaviors~\cite{Lee2004,o2005trust}. Very recently, understanding trust in interaction with ML systems has sparked much interest across disciplines, driven by the rapidly growing adoption of ML technologies. On the one hand, trust in ML systems can be seen as a case for trust in algorithmic systems. Decades of research on this topic yielded complex insights on humans' inclination to trust algorithms and what factors impact the trust. For example, while some studies found an "algorithm aversion" where people stop trusting algorithms after seeing mistakes~\cite{dietvorst2015algorithm}, others found the reverse tendency of "automation bias" with which people overly rely on delegation to algorithms~\cite{Cummings2004}. On the other hand, ML systems present some unique challenges for fostering trust. One is their challenge for scrutablity, especially given the increasing usage of "black box" ML models such as neural networks. Another challenge is their inherent uncertainty, since a ML system can make mistakes in its prediction based on learned patterns, and such uncertainty often cannot be fully captured before deployment using testing methods.

While many emphasized the requisite of transparency for trusting AI~\cite{Doshi-Velez2017,Ribeiro2016}, several recent empirical studies found little evidence that the level of transparency has significant impact on people's willingness to trust a ML system, whether by using a directly interpretable model, allowing user to inspect the model behavior, showing explanation or reducing the number of features presented~\cite{Cheng2019,Poursabzi-Sangdeh2018,kunkel2019let,Schaffer2019}. Many reasons could have contributed to this lack of effect. One is the complex mechanism driving trusting behaviors. According to theories of trust~\cite{HarrisonMcKnight1998,Cummings2004,johnson2004type}, trusting behaviors such as adopting suggestions are not only driven by a more positive perception of the trustee but also other factors such as one's disposition to trust and situation awareness. In fact, several studies suggest that overloading users with information about the system could potentially harm people's situation awareness and lead to worse performance or decision-making outcome~\cite{Cummings2004, Poursabzi-Sangdeh2018,Schaffer2019}.

Perhaps more critically, the premise that transparency or showing information to faithfully reflect the model's behavior should \textit{enhance} trust is questionable, because enhancing trust for an inferior model is deceiving. Instead, in this paper we focus on the goal of \textit{calibrating} trust, to help people correctly distinguish situations to trust or distrust an AI.  While the concept of trust calibration has been studied for automation~\cite{Lee2004, McGuirl2006,Helldin2013,Pop2015}, as to prevent both automation aversion and automation bias, it is not well understood in the context of AI systems. In one relevant study~\cite{Dodge2019}, Dodge et al. compared the effect of different explanation methods for calibrating perceived fairness of ML models, i.e. distinguishing between statistically fair and unfair models. They found that local explanations, by highlighting unfair features used for individual predictions, appear to be more alarming than global explanations when used to explain an unfair model's decisions, and thus more effective in calibrating people's fairness judgment of ML models. Different from Dodge et al., we explore the effect of local explanation on calibrating trust for different predictions made by the same model, instead of calibrating human perception of different models.

As we discussed, calibrating trust for individual predictions is especially important in AI-assisted decision making scenarios. We note several recent studies employed similar AI-assisted decision-making setups and studied how various model related information impacts trust and decision outcome~\cite{Lai2019a,Yu2019,Yin2019,Poursabzi-Sangdeh2018,Schaffer2019}. Multiple studies examined the effect of accuracy information~\cite{Lai2019a,Yu2019,Yin2019}, and found people to increase their trust in the model when high accuracy indicators are displayed, reflected both in subjective reporting and more consistent choices with the model's recommendations. Closest to ours is the work by Lai and Tan~\cite{Lai2019a}, where they studied the effect of showing prediction (in contrast to baseline without AI assistance), accuracy and multiple types of explanation for AI assisted decision-making in a deception-detection scenario. They found that all these features increased people's trust, measured as acceptance of the AI's recommendation as the final decision, and also the decision accuracy. However, a caveat in interpreting the results is that the AI used in this task surpasses human performance by a large margin (87\% compared to 51\%), so any features that manifest the AI's advantage could potentially increase people's willingness to trust the AI, which by default would improve the decision outcome. In fact, observing the results reported by correct versus incorrect model decisions, all these features increased participants willingness to accept the AI's prediction regardless of its correctness, which is evidence that they are ineffective in calibrating trust.

\section{Experiment 1: Effect of showing AI confidence score}
In the first experiment, we tested the following hypotheses with a case study of AI-assisted prediction task:

\begin{itemize}
    \item Hypothesis 1 (\textbf{H1}): Showing AI confidence score improves trust calibration in AI such that people trust the AI more in cases where the AI has higher confidence. 
    

    \item Hypothesis 2 (\textbf{H2}): Showing AI confidence score improves accuracy of AI-assisted predictions.

    
    
\end{itemize}

    H2 is based on the assumption that if H1 holds, then humans may be able to adopt the AI's recommendation at the right time and avoid following wrong recommendations. In addition, we also explored the following research questions:
\begin{itemize}
  \item Research Question 1 (\textbf{RQ1}): How does showing AI's prediction versus not showing, affect trust, accuracy of AI-assisted predictions, and the effect of confidence score on trust calibration? 
  \end{itemize}
  While the former is a common AI-assisted decision-making scenario where the AI gives direct recommendations, the latter represents a scenario where the human has to make blind delegation to the AI without seeing its output. Blind delegation can happen in real-world scenarios where delegation has to happen beforehand, or when the AI decisions have latency. We were also interested in it as a stricter test of trust and trust calibration, following the setup used in Bansal et al.~\cite{bansal2019updates} to test mental modeling of error boundaries.

  \begin{itemize}
    \item Research Question  (\textbf{RQ2}): How does knowing to have more domain knowledge than the AI affect humans' trust, accuracy of AI-assisted predictions, and the effect of confidence score on trust calibration? 
\end{itemize}

To achieve these goals, we designed a prediction task in which participants could  achieve comparable performance to an AI model. This task served as the foundation for both the first and the second experiment.

\subsection{Experimental Design}
\subsubsection{Participants}
We recruited 72 participants from Amazon Mechanical Turk for this first experiment. 19 participants were women, and 2 declined to state their gender. 16 participants were between Age 18 and 29, 32 between Age 30 and 39, 15 between Age 40 and 49, and 9 over Age 50.

\subsubsection{Task and Materials}
\begin{figure}
    \centering
    \includegraphics[width=\columnwidth]{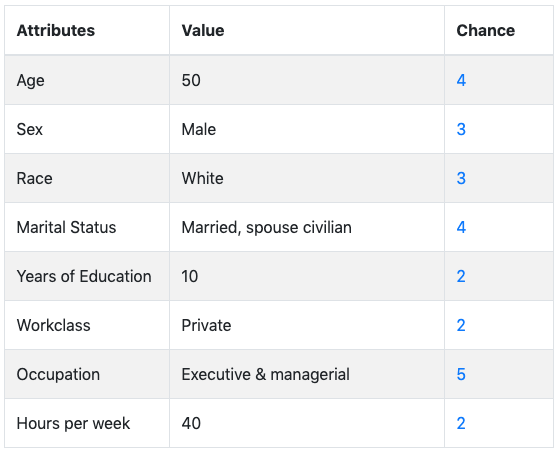}
    \caption{A screenshot of a profile table shown in the experiment. The table lists eight attribute values and their corresponding chances (out of 10) that a person with the same attribute value would have income above \$50K.}
    \label{fig:profile}
\end{figure}

We designed an income prediction task where a participant was asked to predict whether a person's annual income would exceed \$50K based on some demographic and job information. The data used for the task was the 1994 Census Data published as the Adult Data Set in UCI Machine Learning Repository \cite{Dua:2019}. The entire dataset has 48,842 instances of surveyed persons, each described by 14 attributes. These people's annual income, recorded as a binary value indicating above/below \$50K, was used as the ground truth for assessing the participants' prediction accuracy. ML models are trained based on a sample of the dataset to make recommendations to the participants. We selected 8 most important features out of the 14 attributes (as determined by the feature importance values of a Gradient Boosting Decision Tree model over all the data) as features for the models, and as profile features shown to the participants in the prediction trials. The model was trained based on a 70\% random split of the original data set, while the prediction trials given to the participants were drawn from the remaining 30\%. Each prediction trial was shown to the participants with the eight profile attributes in a table like Figure~\ref{fig:profile}.


We intended to create a setup close to real-world AI-assisted decision scenarios where the humans  have comparable domain knowledge with the AI and are motivated to optimize the decision outcome. We took two measures to improve the ecological validity. First, the decision performance was linked to monetary bonus, with a reward of 5 cents if the final prediction was correct and a loss of 2 cents if otherwise (in addition to a base pay of \$3). Prior research showed that such a reward design is effective in motivating participants to optimize the decision outcome~\cite{young1967twelve,bansal2019updates}

Second, since MTurk workers were unlikely familiar with this task, we boosted their domain knowledge and performance by a training task (detailed in Section \ref{precedure}) and an additional piece of information---the third column in Figure~\ref{fig:profile} showing the chance a person with that attribute-value earning income above \$50K on a scale of 0 to 10. This chance number was calculated from the training dataset based on the percentages of people with the corresponding attribute-value earning income above 50K. We multiplied the percentages by 10 and rounded the number since prior work shows that people understand frequencies better than probabilities \cite{Lai2019a}. For example, in Figure~\ref{fig:profile}, the chance value for occupation indicates that 5 people out of 10 with the occupation of Executive \& Managerial have annual income above 50K. For continuous values like Age and Years of Education, chance is calculated over a range, e.g. Age between 45 and 55. The specific range is shown when the participant hovered the mouse pointer on the chance number.

The chance number can be seen as analogous to learning materials that experts may have in real-world scenarios. For example,  decision-makers often have access to statistics of historical events. However, these statistics do not obviate the need for human decision making to synthesize various information. This is also reflected in our task in that the chance values only show probabilities conditioned on single attributes, and the participants still had to learn to combine them to form a prediction based on all attributes.

\subsubsection{Design}
We designed three experimental factors to evaluate the effect of showing confidence scores (\textbf{H1} and \textbf{H2}),  as well as to explore the difference in showing prediction (\textbf{RQ1}) and in scenarios where humans have additional knowledge (\textbf{RQ2}). This 2x2x2 design yields a total of 8 conditions, and we randomly assigned 9 participants to each condition.

\textbf{Show vs. not show AI confidence}.  Studying the effect of confidence scores on people's trust in AI and AI-assisted prediction outcomes is the main goal of this experiment. Confidence is defined as the model's predicted probability for the most likely outcome. For certain ML models, their predicted event probabilities may deviate substantially from the true outcome probabilities (this is called poor calibration). We checked our models and found that their probabilities matched the outcome probabilities very well. Like the chance number, we stated confidence probabilities as frequencies in messages like this: "The model's prediction is correct \textit{N times out of 10} on individuals similar to this one", where N is the rounded number of confidence probability multiplied by 10.

\textbf{Show vs. not show AI prediction}.  We compared a scenario where human had access to the AI's prediction to assist their final decision, versus one where the human had to choose whether to delegate the task to AI without seeing the prediction.  The latter was a stricter test of people's trust and trust calibration. In both conditions, feedback were provided on whether each trial was correct or not, so participants would still experience the AI's performance in conditions where the AI's predictions were not shown.



\textbf{Full vs. partial model}. We explored whether it made a difference when people knew they had access to more information than the AI. This situation is common in real-world AI-assisted decision making, as human experts often posses domain knowledge that is not captured by the data to train the AI. For this purpose, we trained a second partial model without the most important attribute, marital status. Note our focus was not to test human trust on an inferior model, as the accuracy of the partial model (83\%) was only slightly less than that of the full model (84\%) when evaluated on a reserved 20\% test set. Instead, we were interested in the effect of subjectively knowing to have more domain knowledge than the AI on people's trust and decision-making. Therefore, for participants assigned to the partial model condition, we explicitly told them the model was not considering the martial status attribute, and further highlighted the point by distinguishing the marital status attribute in the profile table with a description text "extra information for you".


Since the focus of this research was on calibration of trust for cases where AI prediction was more or less reliable, instead of random sampling, we opted for stratified sampling of cases across different \textbf{confidence levels}. This would increase the number of cases where the AI was less certain about, and allow us to better compare the effect of studied features on cases with different certainty levels. The confidence scores of the model for a binary prediction ranged from 50\% to 100\%. We divided this range into five bins, each covers a 10\% range, and randomly sampled 8 trials from each bin for a participant. The order of these trials was randomized.


\subsubsection{Procedure} \label{precedure}
Upon accepting the task on Amazon Mechanical Turk, participants were brought to our experimental website. They were asked to first give their consent, then read the instruction about the experiment, including the goal of the task and how to read the profile table. The instruction was tailored for the condition the participants were assigned to. 

Next, they were given 20 training trials to practice. In each training trial, after participants gave their predictions, they were shown the actual income category of that person as well as the AI's prediction, so that they could learn from the feedback and assess the AI's accuracy for different cases. They were also shown the AI's confidence level if they were assigned to the with-confidence conditions. After finishing all training trials, participants were told their accuracy and the model's accuracy for the last 10 training trials. 

They then proceeded to the 40 task trials, where they were asked to make their own prediction first. They were then shown the version of AI information (with/without confidence, with/without prediction) depending on which condition they were assigned to. Then the participants were asked to choose their own or the model's prediction as their final prediction. Finally, a feedback message was shown about whether the participant and the model were correct. In the with-prediction conditions, if the participant's own prediction agreed with the AI's prediction, we automatically took that prediction as the final prediction. A 10-second count down was imposed on each trial before the prediction submission button was enabled, encouraging participants to pay more attention in each decision. After the 40 task trials, participants completed a demographic survey.

As discussed, participants received a base pay of \$3 in addition to the performance-based bonus payment (plus 5 cents if correct and minus 2 cents if wrong). On average, each participant received \$1.16 bonus, and a total of \$4.16 compensation for completing the half-hour long experiment.



\subsection{Results}
\subsubsection{Trust}
 Prior work suggests that subjective self-reported trust may not be a reliable indicator for trusting behaviors~\cite{Schaffer2019,kunkel2019let}, which are what ultimately matter in AI-assisted decision tasks. Therefore, following recent studies~\cite{Yin2019,Poursabzi-Sangdeh2018,Lai2019a}, we measured participants' trust in the AI by two behavioral indicators: 

1) \textit{Switch percentage}, the percentage of trials in which the participant decided to use the AI's prediction as their final prediction. In conditions where the AI's prediction was shown, it was the percentage of trials using the AI's prediction among trials where participants and the AI disagreed. In conditions where the AI's prediction was not shown, it was the percentage of trials in which participants chose to delegate the prediction to the AI among all trials. 

2) \textit{Agreement percentage}, the percentage of trials in which the participant's final prediction agreed with the AI's prediction. 

The main difference between the two measures was that in the with-prediction conditions, the agreement percentage would count the trials in which the participant's and the AI's predictions agreed and automatically counted as the final decision; whereas the switch percentage would only consider cases where they disagreed and had to make an intentional act of switching. Therefore, we consider switch percentage to be a stricter measure of trust, even though agreement percentage was used in prior research \cite{Lai2019a}.

\begin{figure}
    \centering
    \includegraphics[width=2.8in]{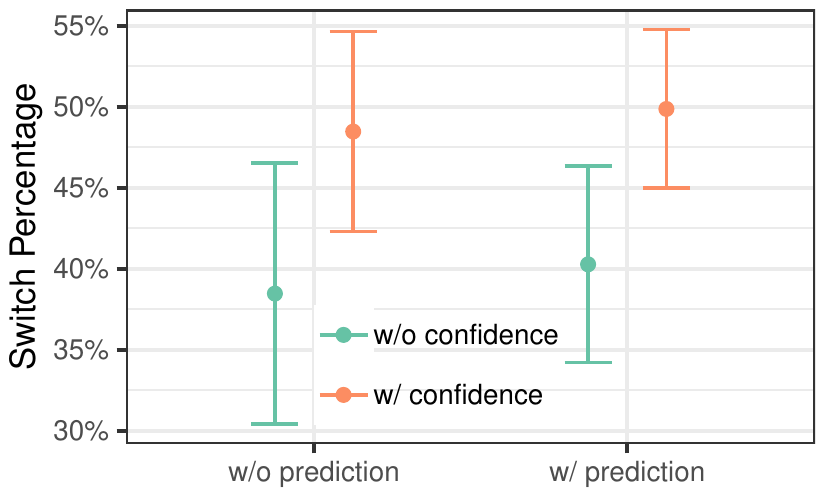}
    \caption{Switch percentage, measured as how often participants chose the AI's predictions as their final predictions, across confidence and prediction conditions. The dots indicate the mean percentages. All error bars in this and subsequent graphs show +/- one standard error.}
    \label{fig:switch}
\end{figure}

Figure~\ref{fig:switch} shows the switch percentage across the prediction and confidence factors. The result that the orange error bars (w/ confidence conditions) are higher than the green error bars (w/o confidence conditions) indicates that the participants switched to the AI's predictions (or decided to use AI in the without-prediction conditions) more often when the AI's confidence scores were displayed. A four factor ANOVA, confidence $\times$ prediction $\times$ model completeness $\times$ model confidence level, confirmed that the main effect of showing confidence scores was significant, $F(1, 64)=4.64$, $p=.035$.

The other two factors, prediction and model completeness, did not have any significant main effect or interaction, partially answering \textbf{RQ1} and \textbf{RQ2}. As can be seen in Figure~\ref{fig:switch}, showing prediction did not affect switch percentages significantly, $F(1, 64)=0.217$, $p=.643$.  The insignificant effect of model completeness, $F(1, 64)=0.07$, $p=.792$, suggests that participants did not distrust the partial model. Given that the two models had similar accuracy, participants acted rationally.

\begin{figure}
    \centering
    \includegraphics[width=2.8in]{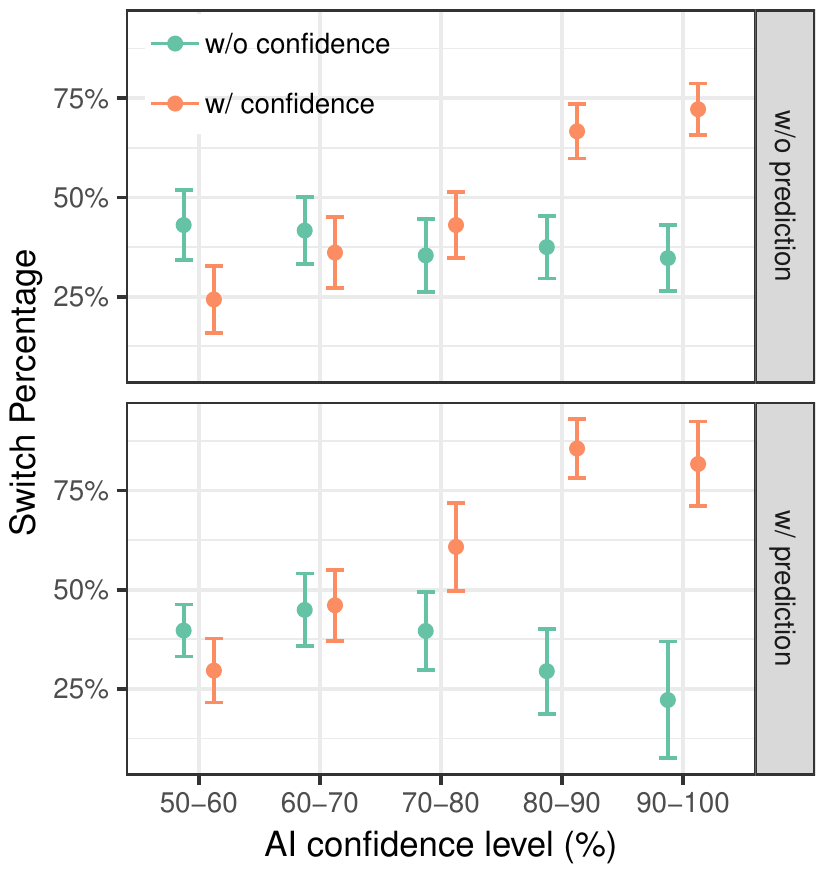}
    \caption{Switch percentage across five confidence levels and various conditions.}
    \label{fig:switch-confidence}
\end{figure}

Figure~\ref{fig:switch-confidence} further examines how showing confidence calibrated trust for cases of different confidence levels. The figure shows that when the AI's confidence level was between 50\% and 80\%, there was not much difference between with- and without-confidence conditions. In fact, participants seemed to trust the model less when AI confidence was shown and was less than 60\%, But when the AI's confidence level was high---above 80\%---participants' trust was significantly enhanced by seeing the confidence scores.  This calibration of trust was confirmed by a statistically significant interaction between showing confidence and the AI's confidence level, $F(4,256)=15.8$, $p < .001$. Further, when the AI confidence score was not shown, participants' trust was generally maintained around the same level across trials of all confidence levels. This was confirmed by an ANOVA on the without-confidence conditions: main effect of confidence level was not significant, $F(4, 128) = 1.84$, $p=.126$.


To answer \textbf{RQ1}, the trust calibration effect by showing confidence score held regardless of whether the model prediction was shown. In other words, high confidence scores encouraged participants to delegate the decision task to the AI even without seeing its predictions. This was confirmed by the insignificant three-way interaction between confidence, prediction, and confidence level, $F(4, 256) = 0.266$,	$p=.899$. 

\begin{figure}
    \centering
    \includegraphics[width=2.8in]{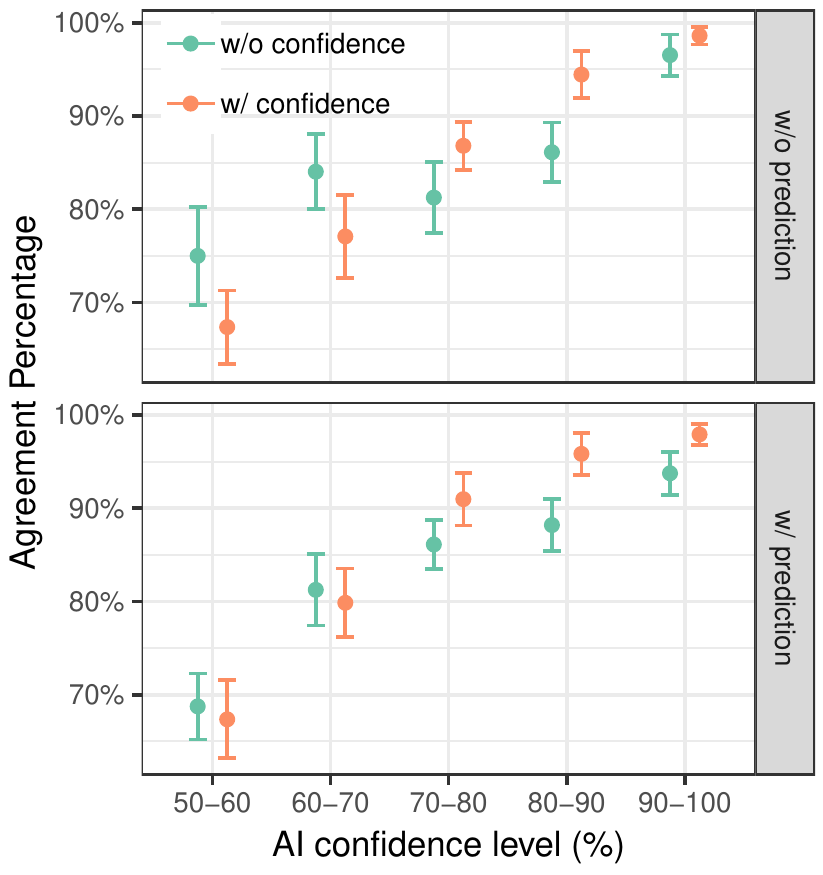}
    \caption{Agreement percentage, measured as how often participants agree with the model's prediction, across confidence levels and various conditions.}
    \label{fig:agreement-confidence}
\end{figure}

A similar pattern was observed in the other trust measure, agreement percentage, as shown in Figure~\ref{fig:agreement-confidence}. When the confidence score was shown, the difference in the agreement percentage between high-confidence levels and low-confidence levels became more pronounced. The calibration effect of confidence score on the agreement percentage, as indicated by the interaction between confidence and confidence levels, was significant, $F(4, 256)=3.82$, $p=.005$. Similarly, this calibration effect held in scenarios of showing and not showing AI prediction, $F(4, 256)=0.331$, $p=.857$. \textbf{H1} was thus fully supported.

\subsubsection{Accuracy}
During the experiment, we collected three types of predictions: (a) participants' own predictions before they saw any information from the AI, (b) the AI's predictions, and (c) the participants' final prediction after seeing AI information, which we call AI-assisted prediction. We measured the accuracy for each type of prediction. On average, the participants' own accuracy was 65\%, with only 14 of 72 participants under 60\%, while the AI accuracy was 75\% (note this number is lower than model accuracy on test data because of stratified sampling for experiment trials). These accuracy numbers did not show statistically significant variations across experimental conditions. Thus, in our task, AI had an advantage over the humans but not by much. This is in contrast to \cite{Lai2019a} where the humans performed substantially worse than the AI (by 37\%).



\begin{figure}
    \centering
    \includegraphics[width=2.8in]{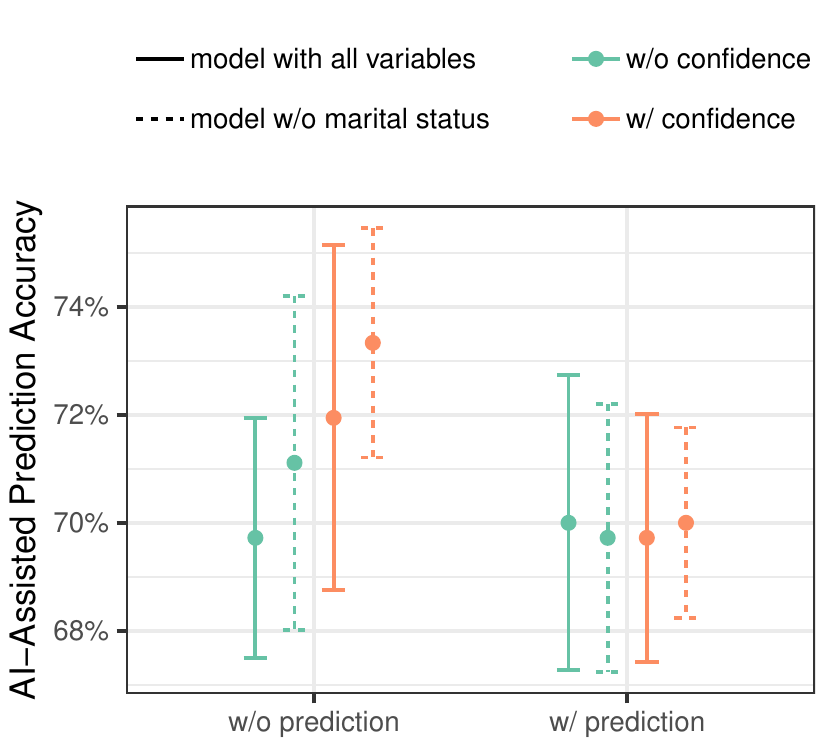}
    \caption{Accuracy of the human and AI-assisted predictions across conditions.}
    \label{fig:human+AI}
\end{figure}

After confirming that displaying confidence both improved overall trust and helped calibrate trust with confidence levels, we investigated whether this translated to improvement in the accuracy of the AI-assisted predictions. Figure~\ref{fig:human+AI} shows this AI-assisted accuracy across conditions. It suggests that there was no significant difference in AI-assisted accuracy across the prediction and confidence conditions. Indeed, an ANOVA showed that only the AI confidence level ($F(4, 256)=79.6$, $p < .001$) and its interaction with model completeness ($F(4, 256)=2.95$, $p = .021$) had significant effect. Furthermore, we also analyzed the difference between AI-assisted accuracy and AI accuracy, and none of the factors showed significant effects. We originally expected that the AI-assisted prediction (i.e. human-AI joint decision) would be more accurate than the AI alone when the AI confidence was low, but that did not turn out to be true.

\begin{figure}
    \centering
    \includegraphics[width=2.8in]{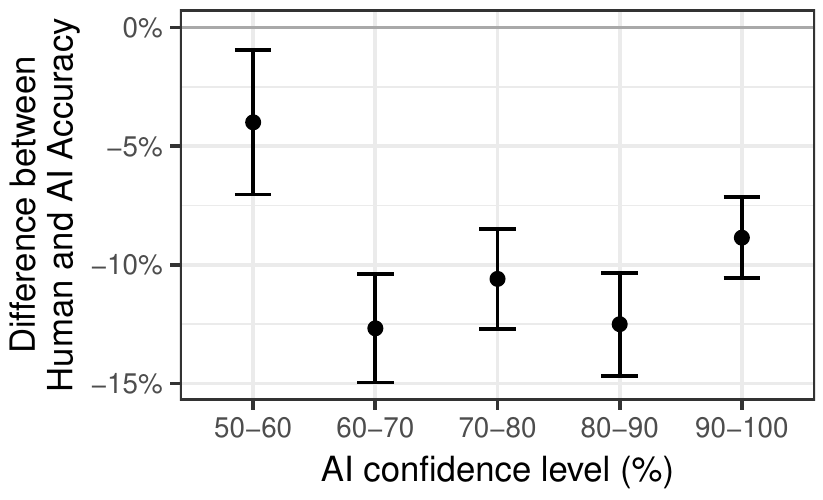}
    \caption{Difference between human and AI accuracy across confidence levels.}
    \label{fig:acc_diff}
\end{figure}

The fact that showing confidence improved trust and trust calibration but failed to improve the AI-assisted accuracy is puzzling, and it rejects our \textbf{H2}. This phenomenon could be explained by the correlation between model decision uncertainty and human decision uncertainty, because trials where the model prediction had low confidence were also more challenging for humans. This can be seen in Figure~\ref{fig:acc_diff} that the humans were less accurate than AI across all confidence levels, although the difference is smaller in the low confidence trials. Therefore, even though showing confidence encouraged participants to trust the AI more in high-confidence zone, the number of trials in which the human and the AI disagreed in these cases were low to begin with, while in the low-confidence zone, human's predictions were not better substitutes for AI's. A caveat to interpret the results here is that if the correlation between human and model uncertainty decreases, for example if the human expert and the model each has a unique set of knowledge, it is possible that better calibration of trust with the model certainty could lead to improved AI-assisted decisions. 


In summary, results of Experiment 1 showed that displaying confidence score improved trust calibration (\textbf{H1} supported) and increased people's willingness to rely on AI's prediction in high-confidence cases. This trust calibration effect held in AI-assisted decision scenarios where the AI's recommendation was shown, and in scenarios where people had to make blind delegation without seeing the AI's recommendation (\textbf{RQ1}). However, in this case study, trust calibration did not translate into improvement in AI-assisted decision outcome (\textbf{H2} rejected), potentially because there was not enough complementary knowledge for people to draw on. While we explored a scenario where participants knew they had additional knowledge that the AI did not have access to, it did not make significant difference in the AI-assisted prediction task (\textbf{RQ2})

\section{Experiment 2: Effect of Local Explanation}
The second experiment examined the effect of local explanations. It had the same setup as Experiment 1, but instead of showing confidence scores, we showed local explanations for each AI prediction. The main hypothesis we wanted to test was that: because local explanation is suggested to help people judge whether to trust a particular prediction~\cite{Ribeiro2016}, and it could potentially expose uncertainty underlying an AI prediction, showing explanation could support trust calibration (\textbf{H3}) and improve AI-assisted predictions (\textbf{H4}).

\subsection{Experiment Setup}
\begin{figure}
    \centering
    \includegraphics[width=2.8in]{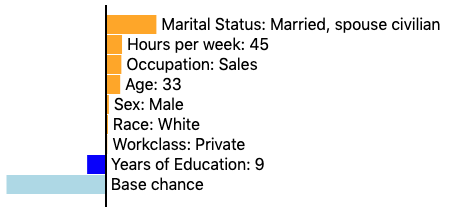}
    \caption{A screenshot of the explanation shown for a particular trial. Participants were told that orange bars indicate that the corresponding attributes suggest higher likelihood of income above 50K, whereas blue bars indicate higher likelihood of income below 50K. The light blue bar at the bottom indicates the base chance---a person with average values in all attributes is unlikely to have income above 50K.}
    \label{fig:explanation}
\end{figure}

We developed a visual explanation feature like the one in Figure~\ref{fig:explanation}.  This visualization explains a particular model prediction by how each attribute contributes to the model's prediction. The contribution values were generated using a state-of-the-art local explanation technique called Shapley method \cite{NIPS2017_7062}.

Experiment 2 was carried out only under the full-model, with-prediction condition. We only tested the full-model condition because the first experiment did not show significant effect of the model completeness. We only tested the with-prediction condition, because even if the prediction was not shown, participants could still derive them from the explanation graphs---if the sum of the orange bars is longer than the sum of the blue bars, the model predicts income above 50K and vice versa. 

Nine participants were recruited for Experiment 2. Four of them were women. One participant was between Age 18-29, four between Age 30 and 39, two between 40 and 49, and two above 50.

\subsection{Results}
The goal of Experiment 2 was to test the effect of local explanation on people's trust in the AI and the AI-assisted decision outcomes, as compared to baseline condition and the effect of confidence scores. Therefore, for the subsequent analysis, we combined the data collected from this experiment with those from the baseline and with-confidence condition of Experiment 1 (all conditions are full-model, with-prediction).

\subsubsection{Trust} 
\begin{figure}
    \centering
    \includegraphics[width=2.8in]{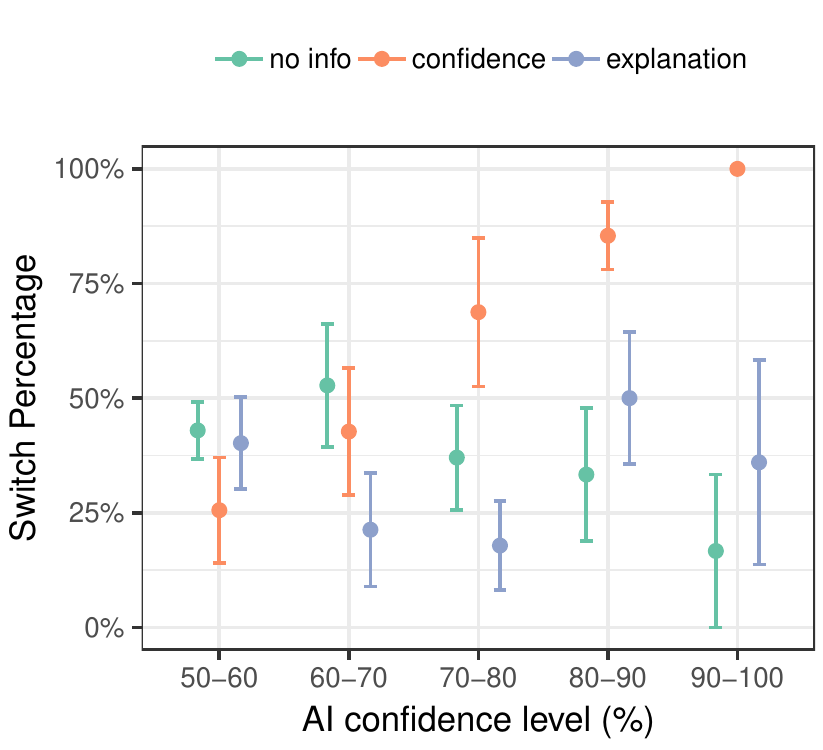}
    \caption{Switch percentage across confidence levels and model information conditions.}
    \label{fig:switch-explanation}
\end{figure}

Figure~\ref{fig:switch-explanation} shows that unlike confidence, explanation did not seem to affect participants' trust in the model predictions across confidence levels. As discussed before, indicated by the orange bars, showing model confidence encouraged participants to trust the model more in high-confidence cases (note that the statistics are not identical to those in Figure~\ref{fig:switch-confidence} because results here only included data in the full-model, with-prediction condition), but the results for explanation (blue error bars) did not show such a pattern. Instead, the switch percentage seemed to stay constant across confidence levels similar to that in the control condition. Results of an ANOVA supported these observations: the model information factor (no info vs. confidence vs. explanation) had a significant effect on the switch percentage, $F(2, 24)=4.17$, $p=.028$, and its interaction with model confidence level was also significant, $F(8, 96)=3.81$, $p < .001$. A Tukey's honestly significant difference (HSD) post-hoc test showed that the switch percentage in the confidence condition was significantly higher than those in the baseline condition (p=.011) and the explanation condition (p < .001), but the explanation condition was not significantly different from the baseline (p = .66).

\begin{figure}
    \centering
    \includegraphics[width=2.8in]{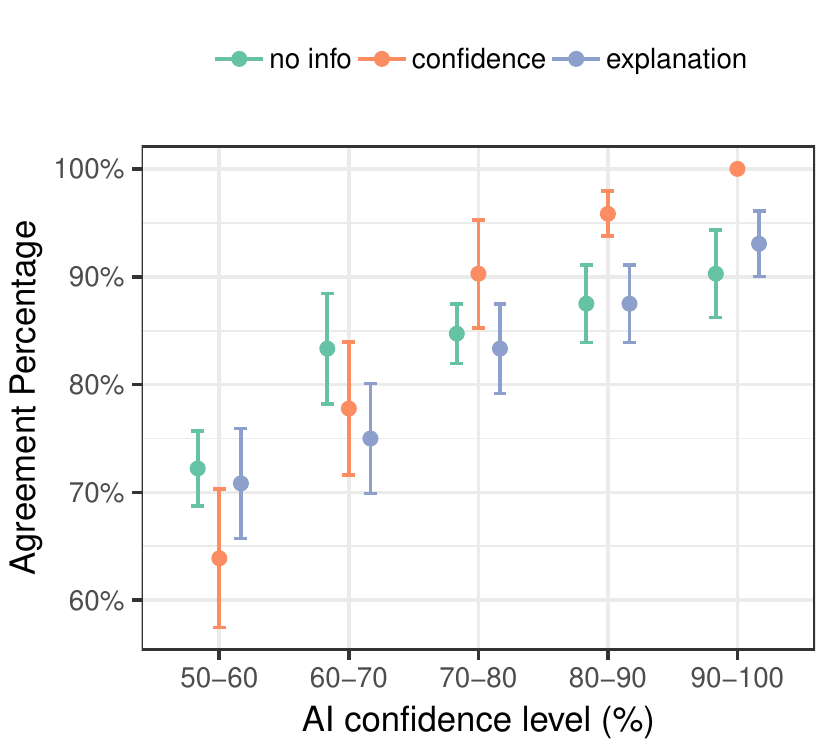}
    \caption{Agreement percentage across confidence levels and model information conditions.}
    \label{fig:agreement-explanation}
\end{figure}

The agreement percentage showed a similar effect, albeit less pronounced. As shown in Figure~\ref{fig:agreement-explanation}, the baseline condition (green) and the explanation condition (blue) had similar agreement percentages, while the with-confidence condition (orange) had higher percentage when the confidence level was above 70\%. Nonetheless, this effect was not significant on this measure, $F(2, 24)=0.637$, $p=.537$. Taken together, \textbf{H3} was rejected as we found no evidence that showing explanation was more effective in trust calibration than the baseline.

\subsubsection{Accuracy} 
\begin{figure}
    \centering
    \includegraphics[width=2.5in]{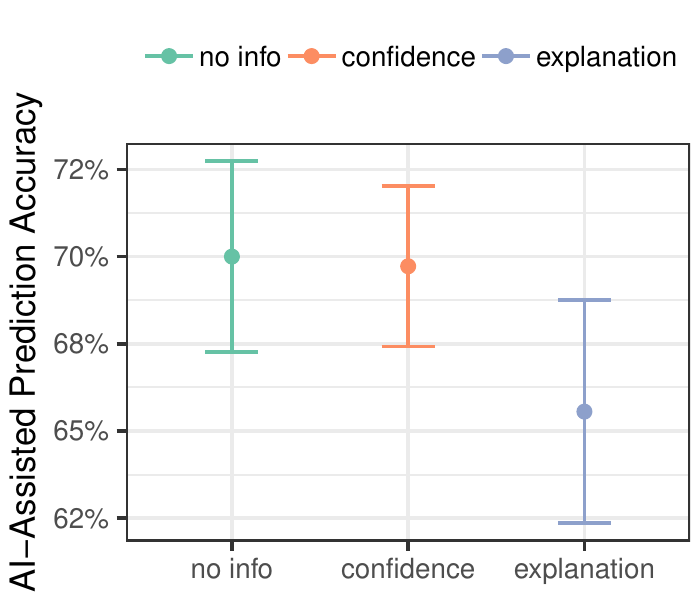}
    \caption{Accuracy of the AI-assisted predictions across conditions.}
    \label{fig:acc-explanation}
\end{figure}

In Experiment 2, the average Human accuracy was 63\%, while the AI's accuracy was again 75\% due to stratified sampling. Figure~\ref{fig:acc-explanation} examines the effect of explanation on the accuracy of AI-assisted predictions. Similar to Experiment 1, we did not find any significant difference in AI-assisted accuracy across model information conditions, $F(2, 24)=0.810$, $p=.457$. If anything, there was a reverse trend of decreasing the AI-assisted accuracy by showing explanation. \textbf{H4} was thus also rejected.

Taken together, the results suggest a lack of effect of local explanations on improving trust calibration and AI-assisted prediction. Our results appeared to contradict conclusions in Lai and Tan's study~\cite{Lai2019a}, which showed that explanations could improve people's trust and the joint decision outcome.  But a closer look at Lai and Tan's results revealed a trend of indiscriminatory increase in trust (willingness to accept) whether the AI made correct or incorrect predictions, suggesting similar conclusion that explanations are ineffective for trust \textit{calibration}. However, since in their study the AI outperformed human by a
large margin, this indiscriminatory increase in trust improved the overall decision outcome. It is also possible in that setup explanations could display the superior capability of the system and more effectively enhance the trust. In the next section, we discuss the implications of differences in the AI-assisted decision task setups and the limitations of local explanations for trust calibration.

\section{Discussions}
We discuss broader implications of this case study for improving AI-assisted decision-making.

\subsection{Mental Model of Error Boundaries}

Consistent with prior work on trust calibration for automation~\cite{McGuirl2006}, we show that case specific confidence information can improve trust calibration in AI-assisted decision making scenarios. In these scenarios, showing confidence is potentially more helpful than showing model-wide information such as accuracy. Bansal et al.~\cite{bansal2019updates} mentioned that well-calibrated confidence scores can potentially help people form a good mental model of AI's error boundaries--understanding of when the AI is likely to err. We recognize that we did not measure people's mental model directly but instead focusing on behavioral manifestation of trust calibration. Developing a good mental model is indeed a higher target, which requires one to construct explicit representation of error boundaries.  With a good mental model, one may be able to more efficiently calibrate trust without the needs to access and comprehend confidence information for every prediction. A recent paper by Hoffman et al.~\cite{hoffman2018metrics} recognizes that forming a good mental model of AI is the key to effectively appropriating trust and usage. The paper also calls out the need to develop methods to measure the soundness of users' mental model, and suggests references from methods in cognitive psychology. Using these methods, future work could examine whether having access to confidence information could effectively foster a mental model of error boundaries.  

However, showing confidence scores has its drawbacks. It is well understood that confidence scores are not always well calibrated in ML classifiers~\cite{Nguyen2015}. Also a numeric score may not be interpreted meaningfully by all people, especially in complex tasks. Moreover, confidence scores alone may be insufficient to foster a good mental model, since it would require people to extract explicit knowledge from repeated experience. Future work could explore techniques to provide more explicit description of error boundaries or low-confidence zones, and study their effect on trust calibration and AI-assisted decision making.

\subsection{Alignment of Human's and AI's Error Boundaries}

Our study found little effect of confidence information on improving AI-assisted decision outcome, even though it improved trust calibration. A potential reason is that, in our setup, the error boundaries of human's and AI's were largely aligned. In other words, in situations where the AI was likely to err, the humans were also likely to err. Participants recruited from Mechanical Turk are not experts in an income prediction task. We attempted to inject domain knowledge by providing participants with chance numbers for each feature, while the model was trained on the same data with the same set of features. While we explored conditions where the human had access to an additional key attribute, it might not have created sufficient advantage for the human. We envision in situations where the AI and the human have complementary error boundaries, trust calibration may be more effective in improving AI-assisted decision outcomes. Future work should test this hypothesis. 

Results of our study show some discrepancies with prior works, especially Lai and Tan's study~\cite{Lai2019a}. We recognize the differences between the setups. While in our study the human and AI had largely aligned knowledge and performance, in~\cite{Lai2019a} the humans had significantly worse performance in the deception detection task. We may consider the setup in~\cite{Lai2019a} to be a situation where the human and the AI not only have unaligned, but also unequal error zones. These comparisons highlight the problem of generalizability from studies of AI-assisted decision making tasks without explicitly characterizing or controlling for the human's performance profile and its difference from the AI's. Our results suggest that such characterization or experimental control may need to go beyond the overall performance, but also consider the alignment of error boundaries between the human and the AI. While how to characterize the level of error boundary alignment poses an open question, we invite the research community to consider it in order to collectively produce unified theories and best practices of AI-assisted decision making.

\subsection{Explainability for Trust Calibration}
Explainable Artificial Intelligence is a rapidly growing research discipline~\cite{Miller2017,Guidotti2018,Carvalho2019,adadi2018peeking}. The quest for explainability has its roots in the growing adoption of high-performance "black-box" AI models, which spurs public concerns about the safety and ethical usage of AI. Given such  "AI aversion", research has largely embraced explainability as a potential cure for enhancing trust and usage of AI. Empirical studies of human-AI interaction also tend to seek validation of trust enhancement by explainability, albeit with highly mixed results. But in practice, there are diverse needs for explainability, as captured by~\cite{Doshi-Velez2017,adadi2018peeking}, including scenarios for ensuring safety in complex tasks, guarding against discrimination, and improving user control of AI. In many of these scenarios, one would desire support for effectively and efficiently identifying errors, uncertainty, and mismatched objectives of AI, instead of being persuaded to over-trust the system. Therefore, we highlight the problem of trust calibration and designed a case study to explore whether a popular local explanation method could support trust calibration.

\begin{figure}
    \centering
    \includegraphics[width=0.49\columnwidth]{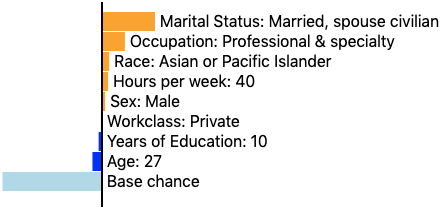}
    \includegraphics[width=0.49\columnwidth]{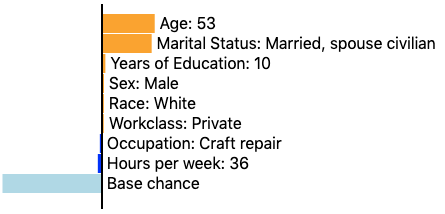}
    \caption{Screenshots of explanation for cases where the model had low confidence.}
    \label{fig:low_exp}
\end{figure}

Unfortunately, we did not find the explanation to create perceivable effect in calibrating trust in AI predictions. This stands in contrast to the findings of \cite{Ribeiro2016} where explanations helped expose a critical flaw in the model (treating snow as Husky), which could help the debugging work. We note the difference in our setup--the classification model may not have obvious flaws in its overall logic and trust calibration may require more than recognizing flaws in the explanation. Figure~\ref{fig:low_exp} lists two examples of explanation shown for a low-confidence prediction. In theory, prediction confidence could be inferred by summing the positive and negative contributions of all attributes. If the sum is close to zero, then the prediction is not made with confidence. However, we speculate that this method of inference might not have been obvious for people without ML training. Instead, one may simply focus on whether the top features and their contributions are sensible. In these two examples, marital status is considered as the main reason for the model to predict higher income. This is a sensible rationale that would frequently appear in explanations regardless of prediction confidence. It is also possible that the explanation created information overload~\cite{Poursabzi-Sangdeh2018} or are simply ignored by some participants. We acknowledge that some of these problems may be specific to the visual design we adopted. It is also possible that the underlying explanation algorithm has its limitation in faithfully reflecting prediction certainty. Nonetheless, our study highlights the importance of studying how an AI explanation design is perceived by a particular group of users, for a particular goal.

There are many other explanation methods and techniques, and it is possible that some are more effective in calibrating trust or exposing model problems. For example, Dodge et al.~\cite{Dodge2019} compared the effect of different explanation methods in exposing discrimination of an unfair model. The study showed that sensitivity based explanation, which highlights only a small number of features that, if changed, could "flip" the model's prediction, is perceived as more alarming and therefore more effective at calibrating fairness judgment than methods that list the contribution of every feature. A study conducted by Cai et al.~\cite{Cai2019} found that comparative explanation, by comparisons with examples in alternative classes, can lead to better discovery of the limitations of the AI, compared to normative explanation that describes examples in the intended class. Although these results imply that some explanation methods may better serve the goal of trust calibration, we know little about the mechanism, neither from the algorithmic side on what makes an explanation technique sensitive to the trustworthiness of a model or prediction, nor from the human perception side on what characteristics of explanations are associated with trust or distrust.

We therefore invite the research community to explore AI explainability specifically for trust calibration, both at the model level and the prediction level. As a starting point, explanation methods and techniques could target a different set of goals in addition to metrics suggested in the current literature such as faithfulness, improved human understanding or acceptance~\cite{Doshi-Velez2017}. For example, explanation that could effectively support trust calibration at the model level should be sensitive to the model performance, while explanation that support trust calibration at the prediction level should be sensitive to the prediction uncertainty. Ultimately, trust resides in human perception and the effect on trust calibration should be evaluated by having targeted users in the loop. Our study provides an example of how to conduct such an evaluation for trust calibration.

\section{Limitations}
One limitation of our study is that our participants are not experts in income prediction. This problem was mitigated by the training task and the access to statistics of the domain (the chance column). The fact that participants' accuracy was only 10\% less than the model trained on a large dataset suggests that these domain-knowledge enhancement measures were effective. Although it is desirable to conduct the experiment with real experts, it can be extremely expensive. Our approach can be considered as "human grounded evaluation" ~\cite{Doshi-Velez2017}, a valid approach by using lay people as "proxy" to understand the general behavioral patterns. 

Another limitation is that we use a contrived prediction task where the participants would not be held responsible. We mitigated the problem by introducing an outcome based bonus reward, which prior studies suggest could effectively motivate optimizing the decision-making. While future study could experiment with scenarios with more significant real-world impact, we note that they have to be executed with caution to avoid ethical concerns.


Lastly, the method that we proposed for calibrating trust---showing model prediction confidence to the decision maker---clearly depends on the model's predicted probabilities being well calibrated to the true outcome probabilities. There are certain machine learning models that do not meet this criterion such as SVM, though this issue can be potentially addressed through Platt Scaling or Isotonic Regression \cite{Niculescu-Mizil:2005:PGP:1102351.1102430}.



\bibliographystyle{ACM-Reference-Format}
\bibliography{acmart}

\end{document}